\documentclass[a4paper,12pt]{article}
\usepackage[utf8]{inputenc}
\usepackage{amsmath}
\usepackage{amsfonts}
\usepackage[numbers]{natbib}
\usepackage{geometry}
\usepackage{tikz}
\title{The Average Patient Fallacy}

\author{
\begin{tabular}{c}
Alaleh Azhir, MD, MSc, Shawn N. Murphy, MD, PhD, Hossein Estiri, PhD
\end{tabular}
\\[6pt]
\small $^{1}$Department of Medicine, Massachusetts General Hospital, Boston, MA, USA\\
\small $^{2}$Department of Neurology, Massachusetts General Hospital, Boston, MA, USA\\}

\begin{document}

\maketitle

\begin{abstract}
Machine learning in medicine is typically optimized for population averages. This frequency-weighted training privileges common presentations and marginalizes rare yet clinically critical cases, a bias we call the average patient fallacy. In mixture models, gradients from rare cases are suppressed by prevalence, creating a direct conflict with precision medicine. Clinical vignettes in oncology, cardiology, and ophthalmology show how this yields missed rare responders, delayed recognition of atypical emergencies, and underperformance on vision-threatening variants. We propose operational fixes: Rare Case Performance Gap, Rare-Case Calibration Error, a prevalence–utility definition of rarity, and clinically weighted objectives that surface ethical priorities. Weight selection should follow structured deliberation. AI in medicine must detect exceptional cases because of their significance.
\end{abstract}

\section{When the ``Average'' Patient is a Fiction}

We begin with a case that, though real, might be dismissed by statistical machines as an outlier. A young, previously healthy twenty-three-year-old develops fulminant Acute Respiratory Distress Syndrome (ARDS) after a short course of Bactrim for acne. She presents with profound hypoxemia, diffuse bilateral infiltrates in her lungs, sterile blood cultures, and an unrevealing autoimmune work-up. Her clinical course progresses rapidly towards multiorgan failure, forcing the care team to consider ECMO and lung transplantation. The clinical team, in discussions with the infectious disease and rheumatology teams, suspects Bactrim lung injury, an exceedingly rare drug reaction. The hospital's Artificial Intelligence early-warning system never issues an alert. It was trained largely on ARDS from sepsis, pneumonia, and trauma, relying heavily on rising inflammatory markers, positive cultures, and stereotyped deterioration patterns. For the model, Bactrim–associated ARDS is statistical noise.

Here, the fallacy becomes visible. We call it: the \emph{average patient fallacy}: the systematic bias toward the statistical mode that diminishes performance for rare but clinically critical cases. For the clinician, rare cases are not outliers to be discarded; they are often the very cases where the stakes are the highest, and the harms from being overlooked are the most severe. To dismiss them is neither scientifically nor ethically excusable.

This article is an exposition of that average patient fallacy: its mathematical origin, its empirical manifestation in clinical practice, and its ethical weight.  We go beyond the threats of imbalanced data\citep{he2009learning,zhou2012ensemble} to specify the fallacy, in the formulation of frequency-weighted optimization. We argue that average patient fallacy is in direct contradiction with the foundational principles of precision medicine.

\section{The Majority-phenotype Machine}

Supervised learning is concerned with the discovery of parameters \(\theta^*\) that minimize an expected loss:

\begin{equation}
\theta^* = \arg\min_{\theta} \mathbb{E}_{(x,y) \sim P} [L(y, f_{\theta}(x))]
\label{eq:objective}
\end{equation}

where \(P\) is the joint distribution of patient features \(x\) and outcomes \(y\), \(L\) is a loss function, and \(f_{\theta}\) is the predictive function. The expectation in this formulation is, by construction, frequency-weighted: common presentations dominate the gradient updates, and the parameters are coerced into a configuration optimized for the majority.\citep{obermeyer2016predicting} While this discussion is focused on supervised learning, it is to be expected that similar prejudices arise in unsupervised or reinforcement learning whenever the optimization is designed to reward frequent patterns over rare but critical ones.\citep{chen2017machine,char2018implementing}

In coarser domains of engineering, this behavior might be defensible as a design specification. In medicine, however, prevalence is a poor proxy for importance. Rare genetic syndromes, atypical drug–response phenotypes, and exceptional therapeutic responders may each occur in less than one percent of cases, yet their significance to both patient care and scientific discovery can be immense.

Here, we are confronted with the collision of two disparate intellectual traditions. Classical machine learning is unapologetically utilitarian: it seeks to maximize overall utility, even at the cost of sacrificing a few points in the tail of a distribution. The moral architecture of medicine is fundamentally different, for it is focused on the irreducible value of the individual case. That difference is not a footnote; it is the very crux of Optimization–Ethics Divergence. \citep{vayena2018machine}

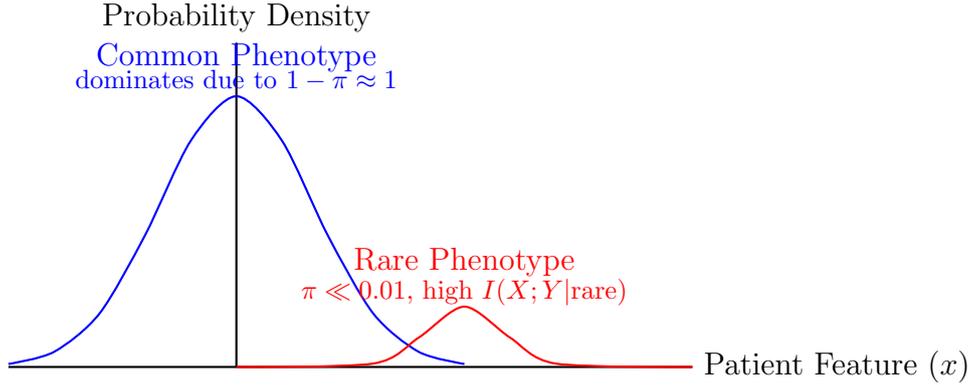
\begin{figure}[ht]
\centering
\begin{tikzpicture}
  \draw[thick] (-3,0) -- (6,0) node[right] {Patient Feature (\(x\))};
  \draw[thick] (0,0) -- (0,4.3) node[above] {Probability Density};
  \draw[blue, thick] plot[smooth] coordinates {
    (-3,0.04) (-2.4,0.20) (-1.8,0.71) (-1.2,1.75) (-0.6,3.00)
    (0,3.59) (0.6,3.00) (1.2,1.75) (1.8,0.71) (2.4,0.20) (3,0.04)
  };
  \draw[red, thick] plot[smooth] coordinates {
    (0,0) (1.8,0.045) (2.4,0.39) (3,0.80) (3.6,0.39) (4.2,0.045) (6,0)
  };
  \node[blue] at (0,4.1) {Common Phenotype};
  \node[blue, align=center, font=\footnotesize] at (0,3.8) {dominates due to \(1 - \pi \approx 1\)};
  \node[red] at (3,1.4) {Rare Phenotype};
  \node[red, align=center, font=\footnotesize] at (3,1.0) {\(\pi \ll 0.01\), high \(I(X; Y | \text{rare})\)};
\end{tikzpicture}
\caption{Mixture distribution of patient phenotypes (Equation \ref{eq:mixture}), showing the dominant common phenotype (blue, weight \(1 - \pi = 0.9\)) and rare phenotype (red, \(\pi = 0.1\) for visibility; in practice, \(\pi \ll 0.01\)). The low weight of rare cases causes their underrepresentation in model optimization (Equation \ref{eq:gradient}) despite their high mutual information (Equation \ref{eq:mutual_info}).}
\label{fig:mixture_distribution}
\end{figure}

Mathematically, this divergence is inevitable. Let us suppose patient features follow a mixture distribution, analogous to a crowded room where only the loudest voices are attended to (see Figure \ref{fig:mixture_distribution}):

\begin{equation}
P(x) = (1 - \pi) \mathcal{N}(\mu_{\text{common}}, \Sigma_{\text{common}}) + \pi \mathcal{N}(\mu_{\text{rare}}, \Sigma_{\text{rare}})
\label{eq:mixture}
\end{equation}

This arrangement mirrors Bayes' theorem, where the posterior probability of a rare phenotype is heavily suppressed by its low prior probability (prevalence \(\pi\)), often leading to its underestimation due to the absence of overwhelming evidence. This is analogous to how frequency weighting discounts rare cases in optimization.

with \(\pi \ll 0.01\) representing the fraction of rare phenotypes. For standard loss functions, the expected gradient contribution from rare cases is necessarily suppressed by their prevalence:

\begin{equation}
\|\mathbb{E}_{\text{rare}}[\nabla_{\theta} L]\| \le \frac{\pi}{1-\pi} \|\mathbb{E}_{\text{common}}[\nabla_{\theta} L]\|
\label{eq:gradient}
\end{equation}

\begin{figure}[ht]
\centering
\begin{tikzpicture}
  \draw[thick] (-2,0) -- (4,0) node[right] {Feature 1: Inflammatory Markers};
  \draw[thick] (0,-2) -- (0,3) node[above] {Feature 2: Visual Acuity};
  \foreach \x/\y in {-1/0.5, -0.5/0, -1.5/1, -0.8/-0.5, -1.2/0.2} {
    \fill[blue] (\x,\y) circle (2pt);
  }
  \foreach \x/\y in {2.5/1.5, 2.8/1.2, 2.2/1.8, 2.6/1.0, 2.4/1.6} {
    \fill[red] (\x,\y) circle (2pt);
  }
  \draw[blue,dashed] (-1,0.25) ellipse (1cm and 1.5cm);
  \draw[red,dashed] (2.5,1.4) ellipse (0.5cm and 0.8cm);
  \draw[<->,thick] (-0.5,0.25) -- (2,1.4) node[midway,above] {Distance};
  \node[align=center, font=\footnotesize] at (-1,-1.5) {Common Cluster\\(\(\theta^*\) optimizes here)};
  \node[align=center, font=\footnotesize] at (2.5,-1.5) {Rare Cluster\\(\(\varepsilon(\pi) \to 0\))};
\end{tikzpicture}
\caption{Feature space showing common and rare clusters (Equations \ref{eq:objective} and \ref{eq:convergence}). Optimization is prejudiced towards the common cluster, marginalizing rare cases. The distance between clusters represents not merely feature-space separation but potential differences in optimal treatment strategy: common diabetic retinopathy may respond to laser photocoagulation, while rare retinal vasculitis requires immunosuppression. The AI may achieve high overall accuracy (e.g., 87.2\% sensitivity for more-than-mild DR) but remains incompetent on rare variants, as suggested by subgroup analyses \citep{abramoff2020pivotal,grzybowski2020ai}.}
\label{fig:feature_space}
\end{figure}
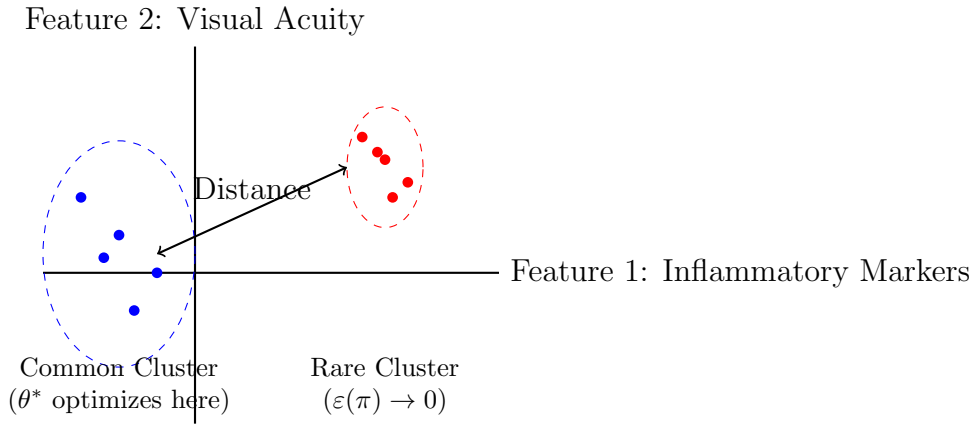

Rare cases, existing as distant outliers in the feature space, are thus systematically ignored by the optimization machinery (see Figure \ref{fig:feature_space}). From an information-theoretic viewpoint, this is a profound misstep, as the mutual information for rare cases may far exceed that for common ones:

\begin{equation}
I(X; Y | \text{rare}) \gg I(X; Y | \text{common})
\label{eq:mutual_info}
\end{equation}

Yet because the optimization weights by frequency, these high-value signals are systematically discounted. As the training converges, the solution approaches:

\begin{equation}
\theta_{\infty} = \theta_{\text{common}} + \varepsilon(\pi)
\label{eq:convergence}
\end{equation}

and the corrective term for rare cases, \(\varepsilon(\pi)\), vanishes as \(\pi \to 0\). In the limit, the model is, by construction, a majority-phenotype machine.

As we proceed, we shall see that this divergence is not a temporary implementation defect, but an inherent consequence of the formalism we have chosen, one whose implications ripple into patient care and impede the progress of personalized medicine.

\section{On Clinical Harm}

The divergence between optimization and ethics does not remain in the algebra. It appears in decisions and harm. The statistical mode exerts a distorting pull. The periphery, where precisely  the rare and critical cases reside, becomes invisible. The net effect is a profile that is logical for the algorithm, but unacceptable for the patient.

Mathematically, this is exactly what one expects from an expectation of $P(x)$. Clinically, one must reject the result. The rare patient is a person whose delayed recognition can separate recovery from relapse, and whose biology may hold clues to therapies for countless others. Ignoring it harms the individual and slows discovery. Consider the following three clinical contexts:

1) In oncology, response distributions are often heavy-tailed. Many patients see a modest benefit from a drug. A few achieve dramatic remission. Those rare responses often carry unique biomarkers, as seen in the discovery of EGFR-mutant non-small-cell lung cancer\citep{Lynch2004}. EGFR mutations occur in approximately 10-15 percent of non-small-cell lung cancers in Western populations, yet response rates to tyrosine kinase inhibitors approach 70 percent in this subgroup compared to 10 percent in unselected patients. Frequency-weighted learning on bulk populations dilutes this signal, impairing the model's ability to identify candidates for targeted therapy. The price of this failure is paid in missed therapies and delayed scientific insight. The fraction of exceptional responders is small, often under ten percent. The yield of studying them, however, can be large. The average patient fallacy wastes that yield.

2) Acute cardiac syndromes present a high-dimensional physiologic space. Common paths include type one myocardial infarction with obstructive coronary disease or decompensated heart failure with congestion, so models trained on large datasets optimize for those dominant patterns. Myocarditis can mimic infarction with chest pain, electrocardiographic changes, and marked troponin release while coronary arteries remain unobstructed and/or heart failure with ventricular dysfunction. Cardiac magnetic resonance or biopsy may be required for its diagnosis.\citep{Yang2021} The rarest, giant cell myocarditis, often declares itself with precipitous biventricular failure, malignant arrhythmias, and mixed shock, where minutes matter for disease-specific immunosuppression and early mechanical circulatory support.\citep{bang2021management} A system tuned to the average infarction or to the average heart failure will perform well there. It will fail here. The cost is avoidable delay and lost survival.

3) In retinal screening, models learn the dominant features of diabetic retinopathy.\citep{grzybowski2020ai}. Rare but vision-threatening variants, such as retinal vasculitis, are displaced in the learned feature space.\citep{ting2019artificial,abramoff2020pivotal} Optimization smooths away their distinctiveness. Early detection becomes less likely for the patients who most require it. The fraction can be five to fifteen percent. The harm from delayed recognition can be irreversible, requiring explicit attention to these subgroups.

\section{Current Mitigation Strategies}

Several techniques exist that purport to address the frequency optimization bias, but each suffers from limitations preventing a complete resolution. While prior approaches, such as focal loss, address the symptoms of imbalance, the average patient fallacy frames the issue as a fundamental defect in the objective.\citep{he2009learning,zhou2012ensemble} This defect must be addressed directly through the formulation of clinically-weighted objectives that embed ethical priorities into the optimization itself. 

\subsection{Relation to Distributionally Robust Optimization}

Our framework shares conceptual affinity with distributionally robust optimization (DRO), which seeks to minimize worst-case loss over a family of distributions rather than optimizing for the empirical average.\citep{duchi2021statistics,sagawa2020distributionally} DRO formulations typically take the form:

\begin{equation}
\theta^* = \arg\min_{\theta} \max_{Q \in \mathcal{U}(P)} \mathbb{E}_{Q}[L(y, f_{\theta}(x))]
\label{eq:dro}
\end{equation}

where $\mathcal{U}(P)$ is an uncertainty set around the empirical distribution $P$. When this uncertainty set is constructed to upweight rare subgroups, DRO can mitigate the average patient fallacy by preventing catastrophic failures on minority populations.

However, DRO and our framework differ in both motivation and implementation. DRO is primarily concerned with robustness to distribution shift—ensuring models perform well when test distributions diverge from training distributions. Our concern is more fundamental: even when distributions are stable, frequency-weighted optimization systematically fails rare but clinically critical cases. DRO's worst-case formulation is also potentially too conservative for clinical deployment, as it may sacrifice substantial performance on common conditions to guard against rare worst-case scenarios.

Our approach can be viewed as a middle path: we explicitly encode clinical priorities through weighting functions rather than implicitly through adversarial uncertainty sets. This makes the value judgments transparent and auditable. Moreover, constrained formulations (developed in Section 6.3) can prevent the "reverse fallacy" of over-optimizing for rare cases at unacceptable cost to common ones—a failure mode that pure worst-case DRO does not inherently guard against.

That said, DRO techniques may prove valuable for implementing our framework, particularly in constructing the uncertainty sets around rare subgroups where data are sparse. The integration of DRO's technical machinery with our ethically-grounded weighting scheme represents a promising direction for future work.

\subsection{Existing Reweighting Techniques}

For example, focal loss modifies the contribution of well-classified examples
\begin{equation}
FL(p_t) = -\alpha_t(1-p_t)^{\gamma}\log(p_t)
\label{eq:focal_loss}
\end{equation}
This improves attention to hard cases and sometimes to rare cases. It is helpful under a moderate imbalance. When prevalence is far below one percent, the basic frequency bias persists. \citep{lin2017focal} 

In another technique, cost-sensitive learning assigns unequal penalties to errors. This requires cost matrices that reflect clinical stakes. Such matrices are often unavailable or are disputed. Static penalties are also too rigid for changing contexts.

These techniques address symptoms rather than the root cause. They modify the optimization locally but do not fundamentally realign the objective function with the moral imperatives of medicine. The average patient fallacy reasserts itself because the underlying frequency-weighted expectation (Equation \ref{eq:objective}) remains uncorrected.

\section{The Precision Medicine Contradiction}

By now, the pattern should be clear. Whether we inspect oncology, critical care, or imaging, the same machinery is at work: frequency-weighted optimization favors the majority phenotype, while rare but high-utility cases are progressively marginalized. The Optimization–Ethics Divergence\citep{vayena2018machine} in its full, crippling manifestation.

On the contrary, the promise of precision (or personalized) medicine is to invert this logic. It rests on the proposition that prevention, diagnosis, and treatment should be tuned to the specific molecular, physiological, and social profile of each patient. Formally, this is about maximizing:

\begin{equation}
T^* = \arg\max_{T} \mathbb{E} [U(\text{outcome}) | x_{\text{individual}}]
\end{equation}

Yet, our AI systems are constructed to optimize:

\begin{equation}
\theta^* = \arg\max_{\theta} \mathbb{E} [U(f_{\theta}(x)) | P(x)]
\end{equation}

This is the \emph{precision–population paradox}: the very tools we hope will enable us to tailor care are optimized for the opposite. The rare cancer responder, the atypical myocarditis patient, the uncommon pattern in a retinal scan—each inhabits a region where small oversights can have large and irreversible consequences.

From the perspective of information geometry, these patients often inhabit regions of steep utility curvature, where small changes in treatment yield disproportionately large differences in outcome. Population-averaged optimization, however, smooths this curvature, making the model less responsive precisely where personalization is most crucial. This is the culmination of the point at which the purported elegance of the mathematics reveals its moral inadequacy \citep{rajkomar2018ensuring,collins2020new,denny2021precision}.

Without explicitly realigning our AI objectives to reflect the values of precision medicine, the average patient fallacy will continue to erode both the scientific promise of AI and the ethical foundation of clinical care. If this erosion is allowed to proceed unchecked, precision medicine risks becoming a promise observed more in rhetoric than in reality.

\section{Toward Measurable and Auditable Solutions}

If the average patient fallacy is to be overcome in practice, its correction must be continuously monitored. Models evolve; they are retrained on new data, they drift as patient populations change, or they are updated in pursuit of gains in aggregate performance. Without explicit, formal checks, the fallacy will quietly reassert itself.

The metrics that make these disparities visible are the operational counterpart to ethical intent. They must act as disciplining constraints, ensuring that rare-case performance is not left to chance or concealed within aggregate statistics. To mitigate the average patient fallacy, technical strategies such as reweighting loss functions, oversampling underrepresented subgroups, or employing fairness-aware algorithms can be applied. These approaches adjust the optimization process, though not without introducing trade-offs, such as reduced overall performance or increased computational cost \citep{hardt2016equality}. Despite these challenges, prioritizing rare cases is consonant with medicine's ethical commitment to all patients. We propose two such metrics:

\subsection{Rare Case Performance Gap (RCPG)}

A direct measure of disparity:

\begin{equation}
\text{RCPG} = P_{\text{common}} - P_{\text{rare}}
\end{equation}

where \(P\) is the performance (e.g., AUROC, sensitivity) in each subgroup. A small RCPG indicates that the model treats the tail of the distribution with the same discipline as the center. A large RCPG is a warning that the precision–population paradox is widening.

\subsection{Rare-Case Calibration Error (RCCE)}

A measure of epistemic honesty:

\begin{equation}
\text{RCCE} = \mathbb{E} [| P(\text{correct} | \text{rare}, \text{confidence}) - \text{confidence} |]
\end{equation}

Poor calibration in rare cases signifies a model that is overconfident where its knowledge is most tenuous—a particularly dangerous failure mode in clinical care, betraying a form of epistemic arrogance. \citep{niculescu2005predicting,wynants2020prediction}

Both RCPG and RCCE can serve as continuous feedback signals, making divergence measurable and hence correctable. They transmute ethical aspiration into operational accountability.

\subsection{Operationalizing the Detection of Rare Cases}

Defining "rare" in practice requires domain-specific thresholds that balance clinical significance with statistical power. We propose a prevalence-utility framework:

\begin{equation}
\text{Rarity Index} = \frac{1}{\text{prevalence}} \times \text{Clinical Utility Score}
\label{eq:rarity_index}
\end{equation}
This index multiplies inverse prevalence by a score reflecting real-world impact, assisting in the prioritization of cases like giant cell myocarditis over common variants.

The Clinical Utility Score must incorporate factors such as mortality impact, therapeutic window sensitivity, and discovery potential. Cases exceeding a defined threshold (e.g., Rarity Index $>$ 100) warrant explicit monitoring.

The calculation of RCPG and RCCE requires maintaining separate performance statistics for rare subgroups, which increases memory overhead by a modest factor (perhaps 2-5x). This cost is negligible compared to the expense of model training.

An overzealous optimization for rare cases can degrade performance on common conditions, creating a "reverse" average patient fallacy. We therefore recommend a constrained optimization:

\begin{equation}
\min_{\theta} \mathbb{E}[L_{\text{common}}] + \lambda \mathbb{E}[L_{\text{rare}}] \quad \text{s.t.} \quad P_{\text{common}} \geq P_{\text{baseline}}
\label{eq:constrained_opt}
\end{equation}
In essence, this balances the improvement on rare cases while constraining the performance on common cases to remain above a pre-specified baseline.

Here, $\lambda$ weights the importance of rare case performance, and the constraint prevents unacceptable degradation for common cases. However, we must confront an uncomfortable truth: the selection of $\lambda$ is not a purely technical matter but a profoundly political one. Who decides that giant cell myocarditis merits $\lambda = 10$ while another rare condition merits $\lambda = 2$? When cardiologists and emergency physicians disagree about the clinical utility of detecting a specific rare presentation, what adjudication process resolves the dispute?

These questions have no algorithmic answer. The determination of $\lambda$ requires deliberative processes involving multiple stakeholders: clinicians with domain expertise, patient representatives, health economists, and ethicists. Different institutions may reasonably arrive at different values based on their patient populations, resources, and mission. A tertiary referral center specializing in rare cardiac diseases might assign higher $\lambda$ values to uncommon myocarditis variants than a community hospital would.

We propose that $\lambda$ values be established through structured clinical consensus processes, analogous to guideline development committees, with explicit documentation of the reasoning and regular reassessment as evidence evolves. The values should be treated as institutional commitments subject to audit and revision rather than as fixed hyperparameters. Critically, the burden of justification should fall on those who would set $\lambda$ near zero—that is, those who would accept systematic neglect of rare cases—rather than on those advocating for higher values.

This procedural approach does not eliminate disagreement, but it makes the value judgments explicit and subject to democratic deliberation rather than concealing them within opaque optimization objectives. The difficulty of calibrating $\lambda$ is not a weakness of our framework; it is an honest acknowledgment of the irreducible value pluralism in medicine.

\subsection{Contextual Optimization}

The relationship between population-level optimization and individual patient care is more complex than a simplistic dichotomy. The practice of medicine operates within resource constraints that necessitate difficult, and sometimes tragic, trade-offs.

In certain contexts, optimizing for the statistical majority is consonant with medical values. Public health interventions like vaccination programs rightfully prioritize common pathogens. Emergency triage systems must rapidly process high-volume, typical presentations. Resource allocation during a pandemic may require utilitarian calculations.

The average patient fallacy becomes intellectually indefensible when rare cases represent: high-stakes missed opportunities (e.g., life-threatening conditions), high discovery potential (e.g., novel therapeutic targets), or matters of health equity (e.g., systematically underserved populations).

Rather than abandoning population-level thinking, we propose a more sophisticated, contextual optimization that adapts to the clinical stakes:

\begin{equation}
\theta^* = \arg\min_{\theta} \mathbb{E}_{(x,y) \sim P} [w(x,y) \cdot L(y, f_{\theta}(x))]
\label{eq:weighted_opt}
\end{equation}
This formulation weights the model's learning by factors of clinical importance, ensuring rare cases are not ignored merely on account of their infrequency.

The weighting function $w(x,y)$ must reflect clinical importance beyond mere frequency:

\begin{equation}
w(x,y) = \text{baseline weight} + \alpha \cdot \text{mortality risk} + \beta \cdot \text{discovery value} + \gamma \cdot \text{equity adjustment}
\label{eq:clinical_weighting}
\end{equation}
This formulation acknowledges resource constraints while ensuring that statistical rarity does not automatically imply clinical insignificance. The challenge is thus transformed from a crude choice between competing ethical frameworks to the disciplined calibration of the weights $\alpha$, $\beta$, and $\gamma$ through clinical consensus and empirical validation.

\textbf{Implementation principle:} The default shall be population optimization for well-understood, resource-intensive scenarios (e.g., routine screening). The shift toward individual-focused optimization is obligatory when the stakes are high and the opportunities for learning are significant.

\section{Conclusion}

The average patient fallacy is not a quirk of machine learning practice; it is a choice of formalism with profound moral consequences. By optimizing for population-averaged performance, we implicitly accept a crude utilitarian calculus: serving the many while systematically failing the few.

Medicine, however, rests on a different foundation. It is grounded in the principle that every patient is owed the best possible care, irrespective of statistical frequency \citep{gillon1985philosophical,jonsen2019abuse,emanuel2020fair}. In the clinic, the "rare" patient is not an abstraction but an individual whose life may depend upon timely, accurate recognition. In research, such patients are often the source of breakthroughs that benefit all.

This moral imperative must, of course, be balanced against practical realities. Resource constraints and the legitimate needs of common conditions cannot be wished away. The path forward lies not in absolutist positions but in the thoughtful calibration of competing priorities, using the formalisms we have outlined.

To continue optimizing solely for the average offers simplicity and impressive headline metrics, but it is an intellectually lazy path. It guarantees that the very patients most in need of individualized care will remain underserved. It also risks erasing the exceptional cases from which medicine has the most to learn.

The practice of medicine has always demanded a disciplined focus on the exceptional case. It is a profound intellectual error to construct algorithms that are, by their very nature, blind to this imperative. To align machine learning with the moral architecture of medicine is not just ethically mandatory; it is measurable, enforceable, and urgent. The choice is not between efficiency and ethics, but between a future where AI strengthens medicine's commitment to all patients and one where it quietly abandons those who need it most.

\bibliographystyle{plainnat}

\end{document}